\pdfoutput=1
\documentclass{article}

\usepackage[final]{acl}
\usepackage{multirow} 
\usepackage{pifont}
\usepackage{times}
\usepackage{latexsym}
\usepackage[utf8]{inputenc} 
\usepackage[T1]{fontenc}    
\usepackage{hyperref}       
\usepackage{url}            
\usepackage{booktabs}       
\usepackage{amsfonts}       
\usepackage{nicefrac}       
\usepackage{microtype}      
\usepackage{xcolor}         
\usepackage{caption}
\usepackage{subcaption}
\usepackage{graphicx}
\usepackage{amsmath}
\usepackage{amssymb}
\usepackage{xspace}
\usepackage[most]{tcolorbox}

\newcommand{\email}[1]{\href{mailto:#1}{\nolinkurl{#1}}}

\definecolor{trc_lightgrey}{RGB}{240, 240, 240}
\definecolor{trc_darkgrey}{RGB}{169, 169, 169}
\definecolor{darkerGrey}{RGB}{105, 105, 105}
\definecolor{trc_lightgreen}{RGB}{245, 250, 241}
\definecolor{trc_darkgreen}{RGB}{207, 228, 186}
\definecolor{trc_lightblue}{RGB}{241, 245, 250}
\definecolor{trc_darkblue}{RGB}{186, 198, 230}
\definecolor{trc_lightred}{RGB}{251, 241, 241}
\definecolor{trc_darkred}{RGB}{236, 185, 191}

\newcommand\codefont[1]
{{\fontfamily{qcr}\selectfont #1}}
\newcommand{\benchmark}{\textsc{LongPiBench}\xspace}

\newtcolorbox{prompt}{
  colback=trc_lightgrey, 
  colframe=trc_darkgrey, 
  colbacktitle=darkerGrey, 
  enhanced,
  boxrule=0pt,
  after skip=0cm,
  before skip=0.3cm,
  right skip=0cm,
  breakable,
  fonttitle=\small\bfseries, 
  fontupper=\small\linespread{1.25}\selectfont, 
  toprule=0pt,
  bottomrule=0pt,
  rightrule=0pt,
  leftrule=4pt,
  arc=0mm,
  skin=enhancedlast jigsaw,
  sharp corners,
  boxed title style={
    frame code={ 
    }
  }
}

\newtcolorbox{prompt_blue}{
  colback=trc_lightblue, 
  colframe=trc_darkblue, 
  colbacktitle=darkerGrey, 
  enhanced,
  boxrule=0pt,
  after skip=0cm,
  before skip=0.3cm,
  right skip=0cm,
  breakable,
  fonttitle=\small\bfseries, 
  fontupper=\small\linespread{1.25}\selectfont, 
  toprule=0pt,
  bottomrule=0pt,
  rightrule=0pt,
  leftrule=4pt,
  arc=0mm,
  skin=enhancedlast jigsaw,
  sharp corners,
  boxed title style={
    frame code={ 
    }
  }
}

\newtcolorbox{prompt_red}{
  colback=trc_lightred, 
  colframe=trc_darkred, 
  colbacktitle=darkerGrey, 
  enhanced,
  boxrule=0pt,
  after skip=0cm,
  before skip=0.3cm,
  right skip=0cm,
  breakable,
  fonttitle=\small\bfseries, 
  fontupper=\small\linespread{1.25}\selectfont, 
  toprule=0pt,
  bottomrule=0pt,
  rightrule=0pt,
  leftrule=4pt,
  arc=0mm,
  skin=enhancedlast jigsaw,
  sharp corners,
  boxed title style={
    frame code={ 
    }
  }
}

\newtcolorbox{prompt_green}{
  colback=trc_lightgreen, 
  colframe=trc_darkgreen, 
  colbacktitle=darkerGrey, 
  enhanced,
  boxrule=0pt,
  after skip=0cm,
  before skip=0.3cm,
  right skip=0cm,
  breakable,
  fonttitle=\small\bfseries, 
  fontupper=\small\linespread{1.25}\selectfont, 
  toprule=0pt,
  bottomrule=0pt,
  rightrule=0pt,
  leftrule=4pt,
  arc=0mm,
  skin=enhancedlast jigsaw,
  sharp corners,
  boxed title style={
    frame code={ 
    }
  }
}

\title{Distance between Relevant Information Pieces  \\[0.2em]  Causes Bias in Long-Context LLMs}

\usepackage{authblk}

\author{

\textbf{
    Runchu Tian$^{1}$\thanks{\ \ Equal Contribution.},
    Yanghao Li$^{1*}$,
    Yuepeng Fu$^{1}$,
    Siyang Deng$^{1}$,
    Qinyu Luo$^{1}$
}
\vspace{-0.8em}
\\

\textbf{
    Cheng Qian$^{1}$,
    Shuo Wang$^{1}$,
    Xin Cong$^{2}$\thanks{\ \ Corresponding Author.},
    Zhong Zhang$^{1}$,
    Yesai Wu$^{1}$
}\\
\textbf{
    Yankai Lin$^{3\dagger}$,
    Huadong Wang$^{1}$,
    Xiaojiang Liu$^{4}$
}\\
$^1$ Department of Computer Science and Technology, Tsinghua University \\
$^2$ Department of Statistics and Data Science, Tsinghua University \\
$^3$ Gaoling School of Artificial Intelligence, Renmin University of China \hspace{0.6em} $^4$Apple Inc. \\

}

\begin{document}

\maketitle

\begingroup
  \renewcommand\thefootnote{}
  \footnotetext{%
  \footnotesize
\hspace*{0.3em} Emails:~trc20@mails.tsinghua.edu.cn,~goody1027@gma\\
\hspace*{2.0em}
il.com,~xin.cong@outlook.com,~yankailin@ruc.edu.cn}
  \addtocounter{footnote}{-1}
\endgroup

\begin{abstract}

Positional bias in large language models (LLMs) hinders their ability to effectively process long inputs.
A prominent example is the "lost in the middle" phenomenon, where LLMs struggle to utilize relevant information situated in the middle of the input.
While prior research primarily focuses on single pieces of relevant information, real-world applications often involve multiple relevant information pieces.
To bridge this gap, we present \benchmark, a benchmark designed to assess positional bias involving multiple pieces of relevant information. It includes various tasks and input lengths. Thorough experiments are conducted with three commercial and six open-source models.
These experiments reveal that while most current models are more robust against the "lost in the middle" issue, there also exist noticeable biases related to the spacing of relevant information pieces.
These findings highlight the importance of evaluating and reducing positional biases for long-context LLMs. Code and data have been made publicly \href{https://github.com/Rachum-thu/LongPiBench}{available}.



\end{abstract}

\section{Introduction}

Large language models (LLMs)~\citep{zhao2023survey, minaee2024large} have made significant progress in various natural language processing tasks~\citep{hendrycks2021measuringmassivemultitasklanguage, han2021pretrainedmodelspastpresent}. In particular, applications such as code repository analysis~\citep{chen2021evaluatinglargelanguagemodels} and information extraction~\citep{kocisky-etal-2018-narrativeqa} often require processing long texts, with context lengths reaching up to 200,000 tokens~\citep{li-etal-2024-loogle, zhang2024bench}. To address these demands, researchers have focused on enhancing LLMs' ability to handle extended inputs effectively~\citep{chen2023longlora, han-etal-2024-lm}. As a result, multiple LLMs have been developed~\citep{dubey2024llama3herdmodels,geminiteam2024gemini15unlockingmultimodal, openai2024gpt4o} which support context lengths of up to one million tokens.

Recent studies have shown that the position of relevant information significantly affects the performance of long-context LLMs~\citep{liu2023lostmiddlelanguagemodels, lei2024s3evalsyntheticscalablesystematic, hsieh2024middlecalibratingpositionalattention}. In "needle in a haystack" tasks, models struggle to utilize information located in the middle of the input, which is known as the "lost in the middle" effect~\citep{liu2023lostmiddlelanguagemodels}. This evaluation method is commonly used to analyze positional bias~\citep{hengle2024multilingualneedlehaystackinvestigating, nelson2024needlehaystackmemorybased}. These analyses~\citep{liu2023lostmiddlelanguagemodels} focused on single relevant information pieces and their positions in the input sequence (front, middle, back), which we refer to as \textbf{absolute positions}.

\begin{figure}
    \centering
    \includegraphics[width=0.8\linewidth]{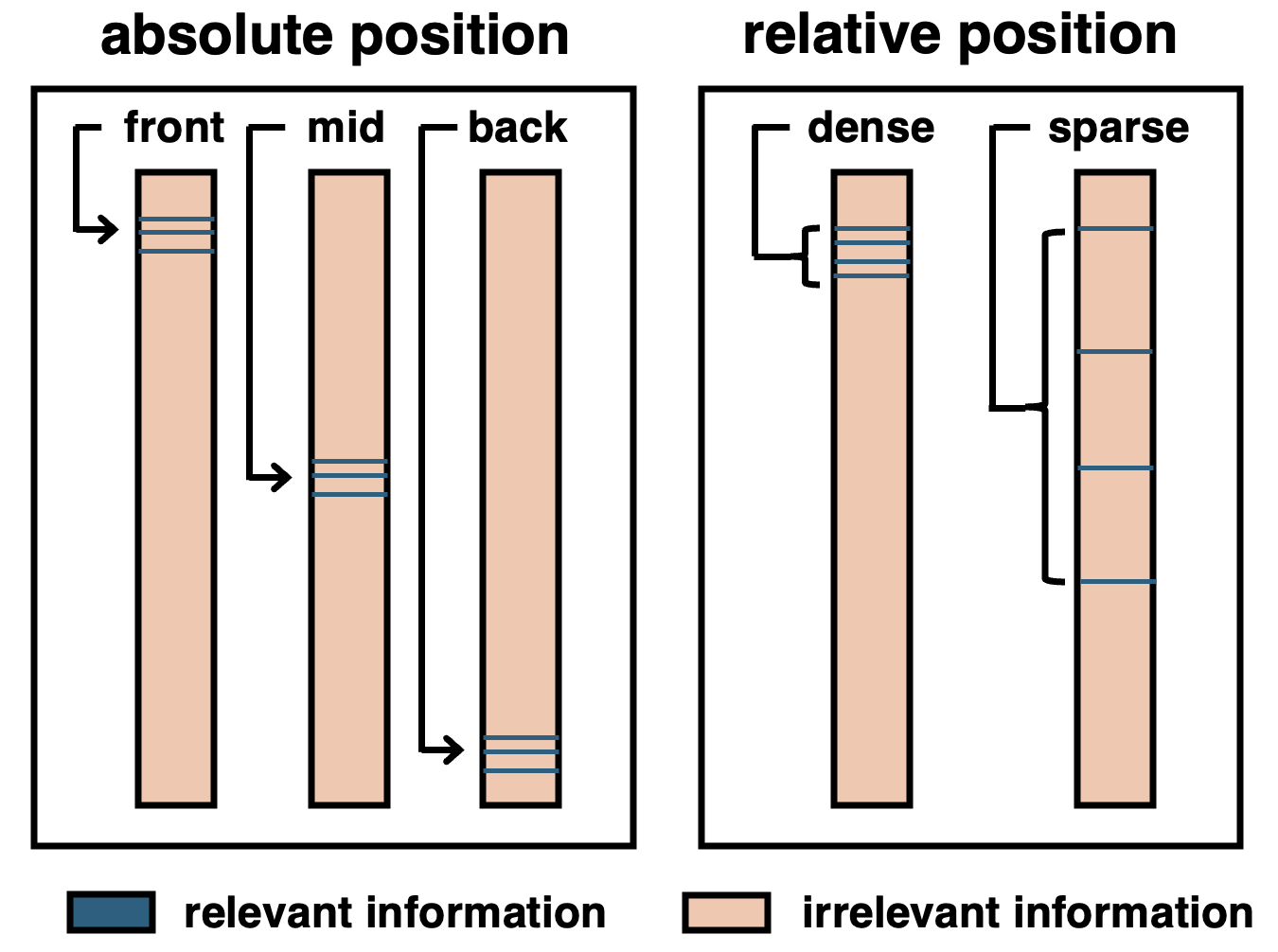}
    \caption{Illustration of absolute position and relative position. Absolute position refers to the location of relevant information within the entire context sequence, while relative position represents the distribution and distance between multiple relevant information pieces.}
    \label{fig:dense-sparse}
    \vspace{-1.5em}
\end{figure}

\begin{figure*}[!ht]
    \centering
    \includegraphics[width=1.0\linewidth, height=0.25\linewidth]{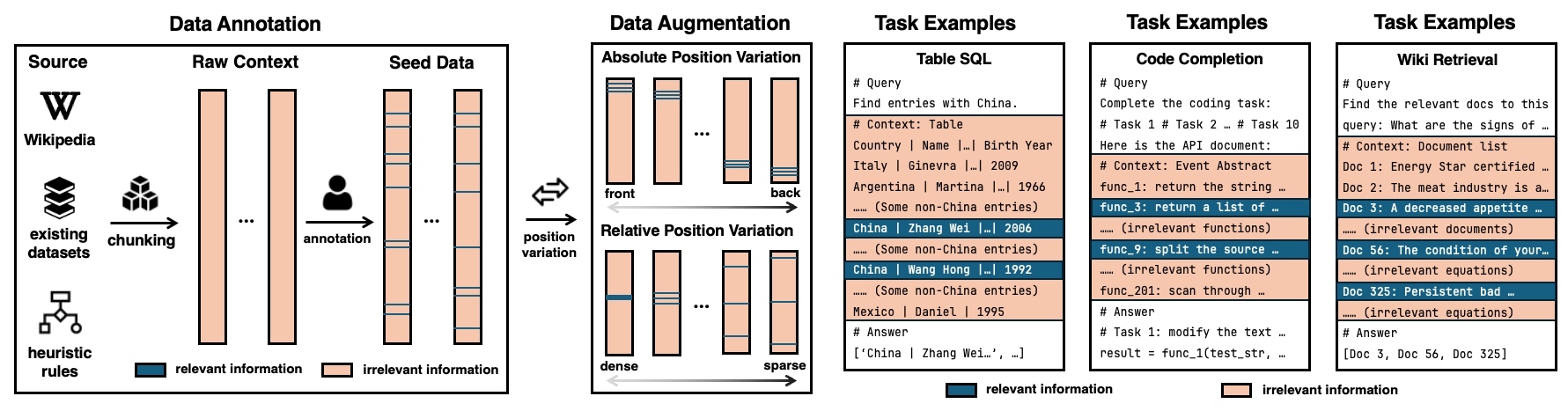}
    \caption{Construction and task examples of \benchmark. We manually annotated seed data and varied the positions of relevant information for data augmentation.}
    \label{fig:construction and example}
    \vspace{-1.2em}
\end{figure*}

 However, real-world tasks like data analysis~\citep{zhang2024bench} often involve multiple pieces of relevant information. This introduces a new characteristic: the distance between relevant information pieces, or how densely they are distributed, which we term as \textbf{relative position}. Evidence from two types of extreme cases indicates that varying relative position may lead to significant bias, impairing LLMs' long-context performance~\citep{lei2024s3evalsyntheticscalablesystematic}. 
However, this kind of biases have not been systematically studied so far, which highlights the need for thorough investigation.


To bridge the gap, we introduce \benchmark, a benchmark designed to evaluate positional bias with multiple relevant pieces. It assesses positional bias in two categories: (1) \textbf{absolute positions}, referring to the location of relevant information within the entire context, and (2) \textbf{relative positions}, referring to the distribution and distance between multiple relevant information pieces. It includes diverse tasks of different complexity and spans four input lengths from 32K to 256K tokens. To the best of our knowledge, \benchmark is the most comprehensive benchmark for isolating and analyzing positional bias in long text models.

We evaluated nine popular LLMs. Our experimental analysis yields several key findings: (1) most current models demonstrate enhanced robustness against "lost in the middle" phenomenon. (2) However, they show biases related to the spacing of relevant information (i.e.\textbf{relative positions}), especially in retrieval tasks. (3) Additionally, we discuss the impact of model size and query-aware contextualization on this issue.

These findings emphasize the importance of evaluating and mitigating positional biases to advance long-context LLM capabilities. 

\section{Related Works}

\subsection{Long-Context Benchmarks}
Many benchmarks have been proposed to evaluate long-context performance of LLMs by designing a variety of tasks with different context length. This field is relatively saturated at present, with some of the representative benchmarks including Long Range Arena~\citep{tay2020long}, Scrolls~\citep{shaham2022scrolls}, ZeroScrolls~\citep{shaham2023zeroscrolls}, Longbench~\cite{bai2023longbench}, L-Eval~\citep{an2023eval}, LV-Eval~\citep{yuan2024lv}, Counting-Stars~\citep{song2024countingstars} and $\infty$Bench~\citep{zhang2024infty}. However, these benchmarks tend to provide only a general conclusion regarding which task types are more challenging, without offering in-depth analysis on positional bias like this paper proposes. 

\subsection{Long Context Data Augmentation}
Data augmentation is a technique widely used~\citep{song2024countingstars, zhang2024infty} in LLM evaluation to expand the datasets for different purposes. Specifically, \citet{levy2024same} explored the impact of input length on reasoning performance using a similar data augmentation approach, adding irrelevant elements to context-relevant elements. While their method shares some similarities with ours, our focus is fundamentally different, leading to entirely distinct conclusions. Their study centers on the overall input length which has nothing to do with positional bias. But we investigate the distance between relevant information pieces, where the input length is fixed. 

\section{\benchmark}

\begin{figure*}[!ht]
    \centering
    \includegraphics[width=0.9\linewidth]{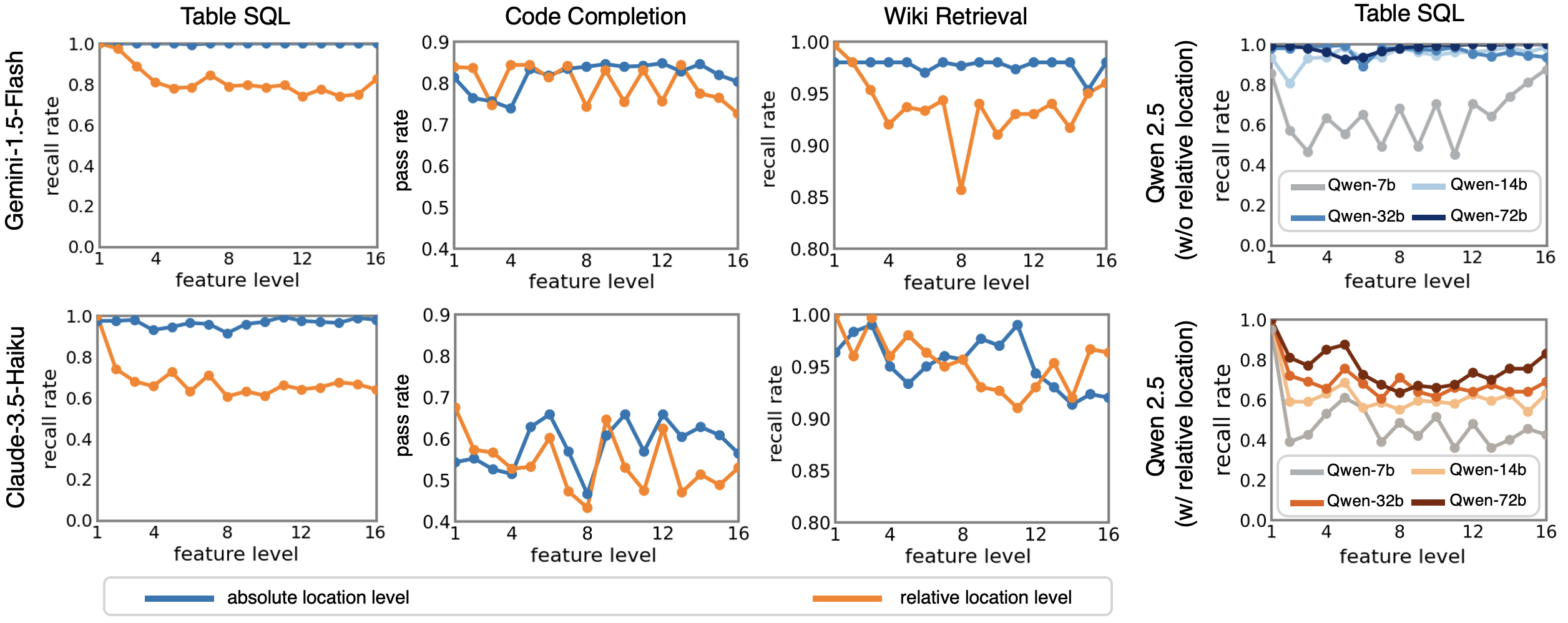}
    \caption{The impact of relevant information's absolute and relative position on Geimini-1.5-Flash~\citep{geminiteam2024gemini15unlockingmultimodal}, Claude-3.5-Haiku~\citep{anthropic2024claude} and Qwen 2.5 model family~\citep{qwen2.5}. A higher absolute position feature level indicates locations closer to the end of input, while a higher relative position feature level indicates a greater distance between relevant pieces of information.}
    \label{fig:main_res}
    \vspace{-1.0em}
\end{figure*}

\benchmark is a dataset designed to evaluate positional bias with multiple relevant information pieces. As shown in Figure~\ref{fig:construction and example}, we first manually annotated several seed examples and then augmented them by varying the positions of relevant information. More details can be found in Appendix~\ref{appendix: benchmark detail}.

\subsection{Core Statistics}
\benchmark contains 3 different tasks, 4 different input length levels\footnote{measured with GPT2Tokenizer~\citep{radford2019language}}: (32$k$, 64$k$, 128$k$, and 256$k$). To analyze the impact of positional bias, we set 16 different absolute and relative location levels respectively. The benchmark is composed of 7,040 instances, each containing around 10 pieces of relevant information. The whole dataset comprises to 845$M$ tokens.

\subsection{Seed Data Annotation}

We manually labeled 15-20 seed data points for three tasks: \textit{Table SQL}, \textit{Code Completion}, and \textit{Wiki Retrieval}, which represent typical use cases in real-world applications of long-context models~\citep{lei2024s3evalsyntheticscalablesystematic, jimenez2024swebenchlanguagemodelsresolve, ajith2024litsearch}. 

\textit{Table SQL} focuses on querying large tables to retrieve entries with specific features, based on experiments from S3Eval~\citep{lei2024s3evalsyntheticscalablesystematic} that explored data with highly variable positions. \textit{Code Completion} requires solving basic programming problems using long masked API documentation. LLMs must identify and apply relevant API details without relying on prior knowledge. The data comes from~\citet{zan2022language}. \textit{Wiki Retrieval} tests LLMs’ ability to find and rerank~\citep{Sun2023IsCG} relevant Wikipedia~\citep{Wikipedia} passages in response to a question, simulating typical information retrieval workflows~\citep{Zhang2023EmpiricalEO}.

Each instance contains 10 relevant pieces of information. This selection was based on an examination of long-context application scenarios, where the number of relevant elements typically falls around the order of magnitude of ten, although it varies across different tasks~\citep{bai2023longbench, wang2024novelqabenchmarkingquestionanswering, dong-etal-2024-bamboo}. Detailed task definitions, examples, and other pertinent details are provided in Appendix~\ref{appendix: benchmark detail}.


\subsection{Data Augmentation}

To analyze the positions of relevant information, we augmented the data by altering the absolute and relative positions of the relevant pieces while keeping all other features unchanged.

We broke down the context into elements based on natural information units: table entries for \textit{Table SQL}, API instances for \textit{Code Completion} and documents for \textit{Wiki Retrieval}. We labeled each element as relevant or irrelevant in a reversal way. We select some elements to be relevant, and then form queries around them, and add irrelevant ones to form the context. By introducing varying amounts of irrelevant information, the context lengths are varied at four levels: 32K, 64K, 128K, and 256K. We then shuffled the element positions to introduce positional variations. Notice that changing the order of elements does not compromise the coherence of the context.

\paragraph{Absolute Position.}

To analyze the impact of absolute position on LLM performance, we manipulated where relevant information appears in the context. Each context was divided into 16 equal segments from start to end. We placed all 10 relevant pieces within a single segment to keep their relative positions consistent. By moving this segment from the first to the last position, we varied the absolute position from the start to the end of the input. The average position of these relevant pieces served as the absolute position metric which is calculated as:
\[
\text{Average Location} = \left( \frac{l - 1}{N - 1} \right) \times L,
\]
where \( l \) is the current level, \( N \) is the total number of levels (16), and \( L \) is the length of the context.

This setup allowed us to assess how model performance changes as relevant information is placed further back in the context.

\paragraph{Relative Position.}

To examine the effect of spacing between relevant information pieces on LLM performance, we created 16 levels of distribution density. Each level represents a different spacing configuration among the 10 relevant pieces. 
At the densest level, all relevant pieces are adjacent with no irrelevant information between them. At the sparsest level, they are evenly distributed throughout the context with equal intervals of irrelevant information. Intermediate levels gradually increase spacing from adjacent to evenly spaced. The distance between each relevant piece is calculated as:
\[
\text{Distance} = \left( \frac{L}{n - 1} \right) \times \left( \frac{l - 1}{N - 1} \right),
\]
where \( n \) is the number of relevant pieces (10), \( l \) is the current level ranging from 1 to \( N \), \( N \) is the total number of levels (16), and \( L \) is the length of the context.

To control for absolute position effects, we randomized the starting position of the first relevant piece in each example. This ensures that any observed performance differences are due to relative spacing rather than absolute positions within the context.

\section{Experimental Setup}
\label{models}

To evaluate the influence of context information positioning on long-text LLMs, we conducted experiments using popular long-context language models.

\paragraph{Models.} We assessed a total of nine LLMs, comprising six open-source and three commercial options. The selection of open-source models includes the 70B model from Llama-3.1-Instruct series~\citep{dubey2024llama3herdmodels}, the 7B, 14B, 32B, 72B models from Qwen-2.5 family~\citep{qwen2.5}, the 8$\times$22B model of WizardLM-2~\citep{xu2023wizardlm}. The commerical models we selected are Gemini-1.5-Flash~\citep{geminiteam2024gemini15unlockingmultimodal}, Claude-3-Haiku~\citep{anthropic2024claude} and GPT-4o-mini~\citep{openai2024gpt4o}. The selected models are good representatives of popular and top-performance long-context models. Due to computational limitations, we evaluated the open-source model only on the \textit{Table SQL} task.


\paragraph{Metric.}
For both the \textit{Table SQL} and \textit{Wiki Retrieval} tasks, performance is measured using recall rate. This metric evaluates the proportion of relevant items included in the output. Formally, given a set of reference items \( D = \{d_1, \dots, d_n\} \) and a set of retrieved/generated items \( \hat{D} \), the recall rate is:  
\[
M_{\text{Recall}} = \frac{|D \cap \hat{D}|}{|D|}.
\]  

In \textit{Table SQL}, \( D \) represents target entries, and \( \hat{D} \) represents the entry present in the output. In \textit{Wiki Retrieval}, \( D \) represents the set of relevant documents, and \( \hat{D} \) represents the top 10 documents retrieved by the model.  

For the \textit{Code Completion} task, performance is evaluated with the pass rate across 8-12 test cases \( T = \{t_1, \dots, t_m\} \). The pass rate is computed as:  
\[
M_{\text{Code}} = \frac{1}{|T|} \sum_{j=1}^{|T|} \mathbf{1}[G \text{ passes } t_j].
\]  
All metrics range from 0.0 to 1.0, where 0.0 means complete failure, and 1.0 means perfect performance.

\paragraph{Context Length.}
Since 32k tokens is the minimal context length supported by tested LLMs, we standardized the context length to 32k\footnote{The minimal context size is 64k, but some tokenizers expand our 64k inputs to nearly 80k, exceeding the limit.} tokens for all experiments.

Detailed discussions on parameter settings and prompt configurations are provided in Appendix~\ref{appendix: experiment setup}.


\section{Results and Discussion} In this section, we analyze the impact of absolute and relative positional bias. And we further analyze these phenomena from two perspectives: the number of parameters and query-aware contextualization. Full Experimental results are available in Appendix~\ref{appendix: experiment results}.

\subsection{Impact of Absolute Position}
As illustrated by the \textcolor[rgb]{0.231,0.459,0.686}{\textbf{blue lines}} in Figure~\ref{fig:main_res}, we progressively shift the interval of relevant information from the beginning to the end.

We observe that (1) some open-source models like Qwen 2.5 (7B)~\citep{qwen2.5} still suffer heavily from the severe "lost in the middle" phenomenon but (2) commercial models and larger open-source models are more robust to the bias of absolute position. Although absolute position still significantly affects the recall rate in the \textit{Code Completion} experiments, this bias becomes less severe in the \textit{Table SQL} and \textit{Wiki Retrieval} tasks.

\subsection{Impact of Relative Position}
As illustrated by the \textcolor[rgb]{0.937,0.525,0.212}{\textbf{orange lines}} in Figure~\ref{fig:main_res}, we progressively increase the distance between relevant pieces of information.

We observe that both open-source and commercial models exhibit noticable biases toward different relative positions. In the case of \textit{Code Completion}, this bias is prominent. As the relative positions of relevant information pieces shift from being fully adjacent to uniformly distributed across the context, the model's performance fluctuates by 20-30\%. For tasks with a stronger retrieval nature, such as Table SQL and \textit{Wiki Retrieval}, the bias even displays certain patterns. Specifically, performance initially declines sharply and then decreases more gradually.

These findings indicate that the relative positioning among multiple relevant pieces of information is a serious and unresolved issue, which may substantially undermine the effectiveness of long-text language models in practical applications.

\subsection{Further Analysis}
\paragraph{Effect of Parameter Size.}


When selecting models for evaluation, we included four variants from the Qwen 2.5 Family~\citep{qwen2.5} with differing parameter sizes. These models exhibit no significant differences in architecture, training methods, or training data. By analyzing their performance under identical positional information features, we can isolate the impact of parameter size on the robustness to positional bias. We use \textit{Table SQL} task, where the pattern is most significant

As illustrated in Figure~\ref{fig:main_res}, for absolute position bias, we found that simply increasing the model parameters from 7B to 14B—while keeping architecture, training methods, and data constant substantially mitigates the "lost in the middle"~\citep{liu2023lostmiddlelanguagemodels} issue. This suggests that robustness to absolute positions may be an "emergent ability"~\citep{wei2022emergent} and increasing the number of parameters can significantly enhances it.

In contrast, regarding biases related to relative positional information, augmenting the number of parameters only yielded minor quantitative improvements and did not alter the pronounced bias trend. This trend remains largely unchanged even in commercial models with approximately hundreds of billions of parameters. These findings indicate that merely increasing parameter size is insufficient to develop robustness to relative positions, and new techniques may be necessary.


\paragraph{Effect of Query-Aware Contextualization.}
\begin{figure}[!t]
    \centering
    \includegraphics[width=0.85\linewidth, height=0.73\linewidth]{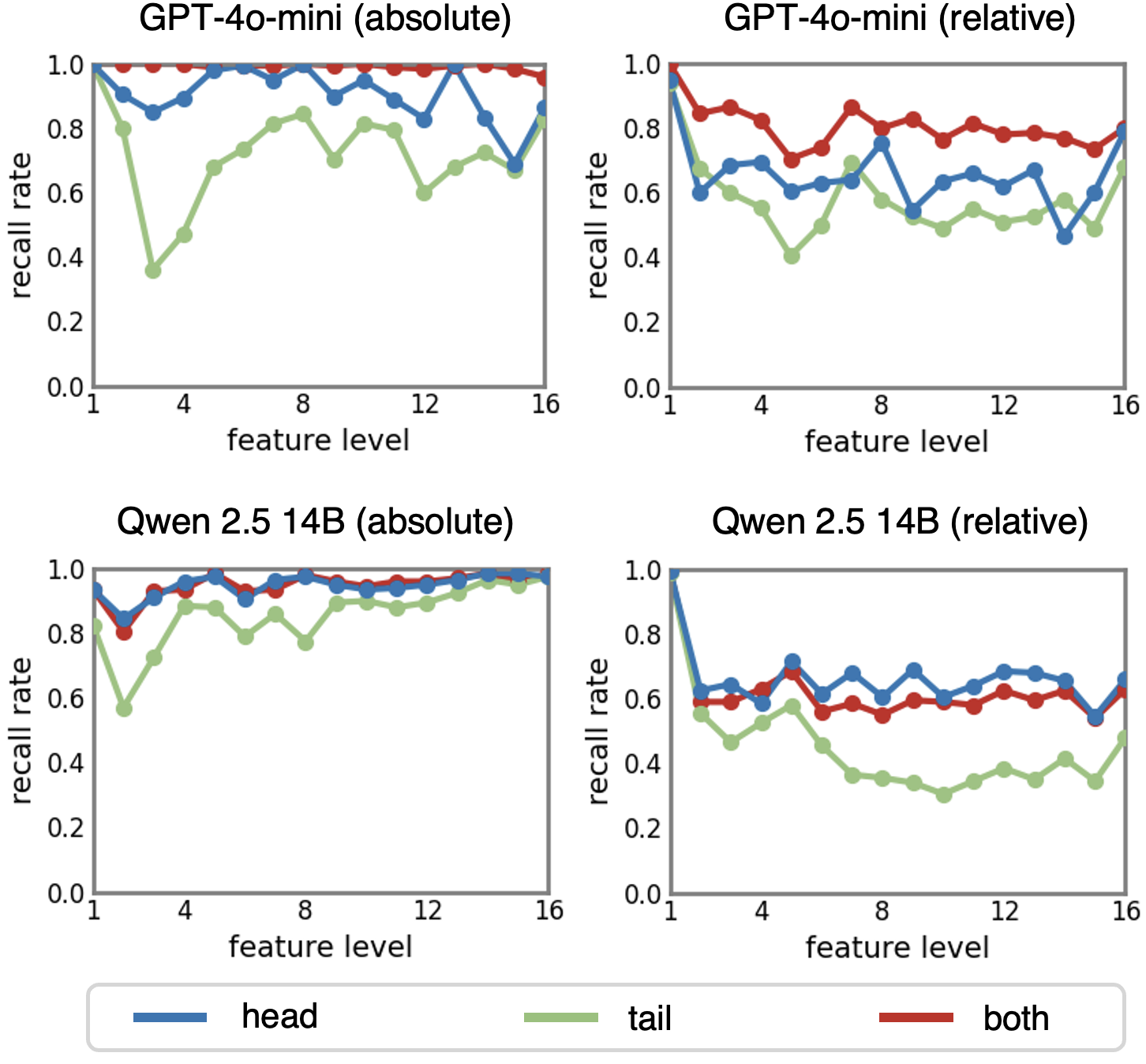}
    \caption{Impact of query placement (beginning, end, both) on the performance of GPT-4o-mini~\citep{openai2024gpt4o} and Qwen-2.5-14B~\citep{qwen2.5} models.}
    \label{fig:query}
    \vspace{-1.6em}
\end{figure}

\citet{liu2023lostmiddlelanguagemodels} demonstrated that the placement of the query (beginning or end of the context) significantly affects the performance of decoder-only models due to unidirectional attention. When the query is placed after the context, the LLM cannot attend to the query token while processing the context tokens. 

As shown in Figure~\ref{fig:query}, our experiments with GPT-4o-mini~\citep{openai2024gpt4o} and Qwen-2.5-14B~\citep{qwen2.5} on \textit{Table SQL} corroborate this observation and confirm that it also holds for bias caused by relative position changes. When the query is placed at the end of the context, the model performs much worse than when the query is at the beginning or both at the beginning and end. However, the difference between placing the query only at the beginning and at both the beginning and end depends on the model. This indicates that for decoder-only long-text models, the position of the query is also crucial in influencing biases related to the absolute and relative positions of relevant information.

\section{Conclusion}

This study investigates a new category of positional bias involving multiple relevant pieces of information in long-context LLMs through three key contributions.

\textbf{(1) Benchmark Development}: We introduce \benchmark, the most comprehensive benchmark for evaluating positional bias in long-text LLMs, assessing both absolute and relative biases.

\textbf{(2) Comprehensive Evaluation}: Using \benchmark, we evaluated nine popular LLMs, investigated the "lost in the middle" phenomenon, and identified novel yet significant biases related to the relative positioning of multiple relevant pieces of information.

\textbf{(3) Findings}: Our experiments show that while LLMs have improved robustness against absolute positional biases, they are still sensitive to relative positional biases, especially for retrieval-intensive tasks. We also explore how model size and query-aware contextualization impact these biases.

These findings emphasize the necessity of continuously mitigating positional biases in long-text models.

\section*{Limitation}

\paragraph{Lack of In-depth Analysis.} Our systematic experiments demonstrate that two types of positional bias exist when multiple related pieces of information are present in the context. We also analyzed how these biases relate to the number of parameters and query contextualization. However, we are currently unable to explain the reasons behind these two positional biases.

\paragraph{Focus on Specific Models.} The evaluation was conducted on a set of nine popular large language models (LLMs), including both open-source and commercial options. However, the findings are limited to these models. The study does not account for the performance of other emerging or less popular models, which might exhibit different results regarding positional biases.

\section*{Ethical Considerations}

\paragraph{Human Annotation.} Our seed construction process involves manual annotation. This annotation was carried out by some of the authors, who are researchers with substantial knowledge in LLM evaluation. Consent was obtained from the individuals whose data we are using or curating. The data collection protocol was approved.

\paragraph{Data Security.} Some data in our Table SQL task may appear to pertain to personal information. However, this data is not actual personal information. Instead, it is generated by us through specific heuristics, eliminating the risk of personal information leakage.

\paragraph{Use of AI assistants} We use GPT-4o~\citep{openai2024gpt4o} for expression modification and grammar sanity check during the composition process.

\section*{Acknowledgment}
This work was supported in part by the National Natural Science Foundation of China (Grant No. 62376273), the Postdoctoral Fellowship Program of CPSF (Grant No. GZB20230343 and Grant No. GZC20240831) and the China Postdoctoral Science Foundation (Grant No. 2023M741945).

We thank all collaborators who made their contributions to this project, including
Runchu Tian (\email{trc20@mails.tsinghua.edu.cn}),
Yanghao Li (\email{goody1027@gmail.com}),
Yuepeng Fu (\email{fuyp22@mails.tsinghua.edu.cn}),
Siyang Deng (\email{sydeng@bupt.edu.cn}),
Qinyu Luo (\email{qinyu_luo@163.com}),
Cheng Qian (\email{qianc20@mails.tsinghua.edu.cn}),
Shuo Wang (\email{wangshuo.thu@gmail.com}),
Xin Cong (\email{xin.cong@outlook.com}),
Zhong Zhang (\email{zhongzhang@tsinghua.edu.cn}),
Yesai Wu (\email{wuyesai@gmail.com}),
Yankai Lin (\email{yankailin@ruc.edu.cn}),
Huadong Wang (\email{huadw2012@163.com}),
and Xiaojiang Liu (\email{xiaojiang_liu@apple.com}).
For further information or collaboration, feel free to contact them.

\bibliography{ref}

\clearpage

\appendix


\section{Details of \benchmark}
\label{appendix: benchmark detail}

\subsection{Task Definitions}
\paragraph{Table SQL}
This task involves retrieving entries containing specific features from a table with a large number of entries. The prototype of this task is primarily derived from experiments in S3Eval~\citep{lei2024s3evalsyntheticscalablesystematic}, specifically those examining information distributions with extreme positional variability.

\paragraph{Code Completion}
This task involves performing basic programming assignments based on the definitions, signatures, examples, and other information provided in API documentation. The task is considered more challenging than Table SQL tasks because an LLM must not only identify which parts of the API documentation are relevant but also correctly utilize them during coding. The data we use originates from the Private Coding Dataset introduced by \citet{zan2022language}. To ensure that the LLM does not rely on internal knowledge about common Python libraries, both the API documentation and task function names have been masked. This privatization process is crucial for evaluating performance on long-text scenarios, as it compels the LLM to extract relevant information directly from the provided context.

\paragraph{Wiki Retrieval}

This task involves identifying relevant passages from Wikipedia~\citep{Wikipedia} pages based on a given question. It is a common scenario in which LLMs are used to rerank relevant passages retrieved through information retrieval systems~\citep{ajith2024litsearch}.

\subsection{Task Examples}

Here are some examples of the three tasks in \benchmark. Queries are placed both before and after the context for better query contextualization. 

\subsubsection{Table SQL}

\begin{prompt}
\textbf{Input}
You are given a table of entries with the following columns: Country, Name, Birth Year, Birth Month, Blood Type. Your task is to find all the entry with the following Country: China. You should return all the entries that match the query as a python list. For example, ['| China | Hong Liang | 1991 | August | A |', ...]. You should not generate anything else. Here is the table:\\
| Country | Name | Birth Year | ... | Blood Type |\\
| Italy | Ginevra | 2009 | February | O |\\
| Argentina | Martina | 1966 | March | B |\\
| Egypt | Salma | 1985 | July | B |\\
...\\
| China | Zhang Wei | 2006 | November | O |\\
...\\
| China | Wang Wei | 1966 | February | AB |\\
...\\
| Australia | Emily | 1983 | December | O |\\
| Italy | Leonardo | 1985 | November | O |\\\\
You are given a table of entries with the following columns: Country, Name, Birth Year, Birth Month, Blood Type. Your task is to find all the entry with the following Country: China. You should return all the entries that match the query as a python list. For example, ['| China | Hong Liang | 1991 | August | A |', ...]. You should not generate anything else.
\end{prompt}
\begin{prompt_green}
\textbf{Ground Truth}

[\\
    "| China | Zhu Wei | 1992 | September | B |",\\
    "| China | Zhang Wei | 1955 | March | O |",\\
    "| China | Zhang Wei | 2006 | November | O |",\\
    "| China | Wang Wei | 2001 | September | B |",\\
    "| China | Yang Wei | 2016 | November | AB |",\\
    "| China | Li Na | 1974 | January | B |",\\
    "| China | Liu Wei | 1975 | November | O |",\\
    "| China | Gao Wei | 1954 | August | B |",\\
    "| China | Zhu Wei | 1989 | September | AB |",\\
    "| China | Wang Wei | 1966 | February | AB |"\\
        ],
\end{prompt_green}

\subsubsection{Code Completion}

Notice that in the \textit{Code Completion} task, the ground truth is provided in its unmasked form, while the LLMs generate code based on the masked API documentation, resulting in masked code as output.
\begin{prompt}
\textbf{Input} 
Please complete the code snippet above according to the provided code snippet and the api doc.

\codefont{
\# Text where substitution will take place

text = 'Thelib\_2 alib\_2 123 apples and 456 oranges.'

\# Define pattern and replacement for substitution

sub\_pattern = r'\\d+'
lib\_2placement = 'NUM'

\# Task 1: Substitute matching text using `sub\_pattern` and `lib\_2placement`

lib\_2sult\_1 = 
print(lib\_2sult\_1)

\# Task 2: ...
}

The following context is a code snippet with the detailed api doc.

\begin{verbatim}
{
  "api_path": "lib_2.submodule_26",
  "api_doc": "Returns complex...",
  "api_signature": "",
  "api_parameters": "",
  "api_parameters_number": "=0",
  "api_returns": ""
},
...
\end{verbatim}
(more API instances)

Please complete the code snippet above according to the provided code snippet and the api doc.

\codefont{
\# Text where substitution will take place

text = 'Thelib\_2 alib\_2 123 apples and 456 oranges.'

\# Define pattern and replacement for substitution

sub\_pattern = r'\\d+'
lib\_2placement = 'NUM'

\# Task 1: Substitute matching text using `sub\_pattern` and `lib\_2placement`

lib\_2sult\_1 = 
print(lib\_2sult\_1)

\# Task 2: ...
}
\end{prompt}
\begin{prompt_green}
\textbf{Ground Truth}
\begin{verbatim}
import re

text = 'There are 123 apples 
and 456 oranges.'
sub_pattern = r'\d+'
replacement = 'NUM'

## task 1
result_1 = re.sub(sub_pattern, 
replacement, text)
print(result_1)
...
\end{verbatim}

\end{prompt_green}

\subsubsection{Wiki Retrieval}
\begin{prompt}
\textbf{Input} Please find the top-10 most helpful Docs that will help answer the question. (You do not need to answer it.)

What are ten easy eco-friendly practices that individuals can adopt in their daily lives?

Here is the context

Doc 1

Gaetano \"James\" Senese (born 6 January 1945) is an Italian saxophonist, composer, and singer-songwriter.  Life and career Senese was born in Naples, the son of Anna Senese and James Smith, an American soldier from North Carolina in Italy because of World War II. Senese's father moved back to the US eighteen months after Gaetano's birth and never returned. Senese started playing the saxophone at 12 years old.

Doc 2

He made his professional debut in the 1960s, as a member of the rhythm and blues band The Showmen (later known as Showmen 2), with whom he won the 1968 edition of Cantagiro. In 1974 Senese co-founded and led the critically acclaimed jazz-rock group Napoli Centrale. After the group disbanded in 1978, he started a long collaboration with Pino Daniele, both in studio and on stage. His first solo album was released in 1983 by Polydor Records.

......

Doc 1128

Release and critical reception Generations in Song was first released on Coldwater Records in 2001. It was originally offered as a compact disc and contained 19 tracks in its original release. On February 10, 2004, the album was re-released on Slewfoot Records in a compact disc format again. However, only 12 tracks were included on the re-release. The album cover was also changed for the re-release of the project.

Please find the top-10 most helpful Docs that will help answer the question. (You do not need to answer it.)

What are ten easy eco-friendly practices that individuals can adopt in their daily lives?

You should output a python list of the Doc Index like ```['Doc 1', ...]``` as your answer
\end{prompt}
\begin{prompt_green}
\textbf{Ground Truth} 

[
            "Doc 920",
            "Doc 927",
            "Doc 935",
            "Doc 942",
            "Doc 949",
            "Doc 957",
            "Doc 964",
            "Doc 971",
            "Doc 979",
            "Doc 986"
]
\end{prompt_green}

\section{Details of Experimental Setup}
\label{appendix: experiment setup}

\subsection{Inference Parameters}

To ensure consistency and reproducibility in our experiments, we standardized the inference parameters across all models during the inference phase. Specifically, we set the temperature parameter (\( \text{temp} \)) to 0.1 and the top-p sampling parameter (\( \text{top}_p \)) to 0.9. This unification of inference settings facilitates the replication of experiments and establishes a consistent evaluation standard across different models.

\subsection{Prompt Template}

For the three tasks, we used the following prompt templates respectively. Notice that we place queries both before and after the context body for better query contextualization.

\subsubsection{Table SQL}

\begin{prompt}
\textbf{Input}
You are given a table of entries with the following columns: Country, Name, Birth Year, Birth Month, Blood Type. Your task is to find all the entry with the following Country: \codefont{\{country\}}. You should return all the entries that match the query as a python list. For example, ['| China | Hong Liang | 1991 | August | A |', ...]. You should not generate anything else. Here is the table:\\
\codefont{\{context\}}\\
You are given a table of entries with the following columns: Country, Name, Birth Year, Birth Month, Blood Type. Your task is to find all the entry with the following Country: \codefont{\{country\}}. You should return all the entries that match the query as a python list. For example, ['| China | Hong Liang | 1991 | August | A |', ...]. You should not generate anything else.
\end{prompt}

\subsubsection{Code Completion}

\begin{prompt}
\textbf{Input} Please complete the code snippet above according to the provided code snippet and the api doc.

\codefont{\{query\}}

The following context is a code snippet with the detailed api doc. 

\codefont{\{context\}}

Please complete the code snippet above according to the provided code snippet and the api doc.

\codefont{\{query\}}
\end{prompt}

\subsubsection{Wiki Retrieval}
\begin{prompt}

\textbf{Input} Please find the top-10 most helpful Docs that will help answer the question. (You do not need to answer it.)

\codefont{\{query\}}

Here is the context

\codefont{\{context\}}

Please find the top-10 most helpful Docs that will help answer the question. (You do not need to answer it.)

\codefont{\{query\}}

You should output a python list of the Doc Index like ```['Doc 1', ...]``` as your answer

\end{prompt}

\section{Details of Experimental Results}
In the main text, for better readability, we only presented the experimental results of a subset of tested LLMs in the form of line charts. Here we present all the experimental results in both tabular and chart form. This will better facilitate the precise display of the experimental results.

\begin{figure*}[!ht]
    \centering
    \includegraphics[width=0.5\linewidth]{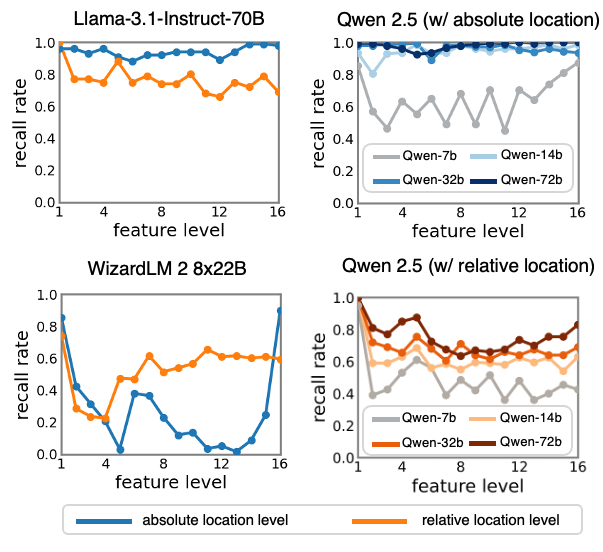}
    \caption{The impact of relevant information’s absolute and relative position for all open-source commercial models. A higher absolute position
feature level indicates locations closer to the end of input, while a higher relative position feature level indicates a
greater distance between relevant pieces of information.}
    \label{fig:full_open}
\end{figure*}

\begin{figure*}[!ht]
    \centering
    \includegraphics[width=0.7\linewidth]{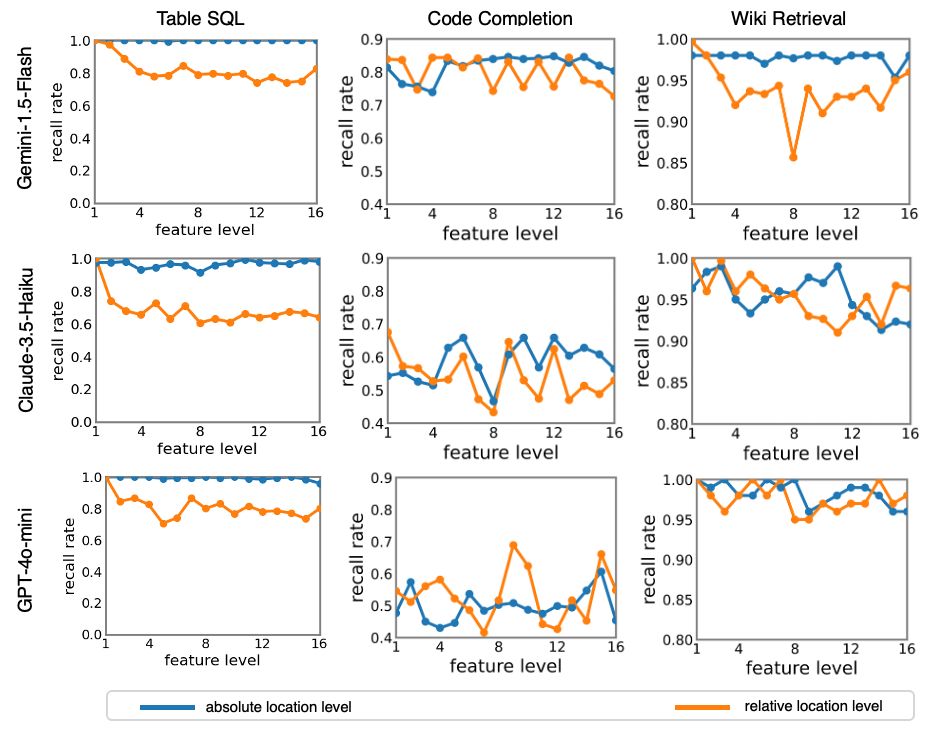}
    \caption{The impact of relevant information’s absolute and relative position for all tested commercial models. A higher absolute position
feature level indicates locations closer to the end of input, while a higher relative position feature level indicates a
greater distance between relevant pieces of information.}
    \label{fig:full_close}
\end{figure*}

Figure~\ref{fig:full_open} and~\ref{fig:full_close} use line charts to illustrate the performance of all selected closed-source and open-source models across the respective test tasks.

Table~\ref{tab:abs} and Table~\ref{tab:rlt} summarize the performance of all models on the \textit{Table SQL} task across different absolute and relative positions. Similarly, Table~\ref{tab:absolute_performance2} and Table~\ref{tab:relative_performance2} present the results for the \textit{Code Completion} task, while Table~\ref{tab:wiki_absolute_performance} and Table~\ref{tab:wiki_relative_performance} correspond to the \textit{Wiki Retrieval} task. Finally, Table~\ref{tab:query_abs} and Table~\ref{tab:query_rlt} show the impact of query contextualization.

\label{appendix: experiment results}

\begin{table*}[htbp]
\centering
\small
\scalebox{0.78}{
\begin{tabular}{lcccccccccccccccc}
\toprule[1.1pt]
\multirow{2}{*}{\vspace{-0.65em}\textbf{Model}} & \multicolumn{16}{c}{\textbf{Performance / \%}} \\
\cmidrule(lr){2-17}
 & Lv. 1 & Lv. 2 & Lv. 3 & Lv. 4 & Lv. 5 & Lv. 6 & Lv. 7 & Lv. 8 & Lv. 9 & Lv. 10 & Lv. 11 & Lv. 12 & Lv. 13 & Lv. 14 & Lv. 15 & Lv. 16 \\
\midrule[0.7pt]
Claude     & 97.5 & 97.5 & 98.0 & 93.0 & 94.5 & 96.5 & 96.0 & 91.5 & 96.0 & 97.0 & 99.5 & 97.5 & 97.0 & 96.5 & 99.0 & 98.0 \\
Deepseek   & 100.0 & 99.5 & 99.5 & 97.0 & 98.5 & 98.5 & 97.5 & 99.5 & 99.0 & 95.5 & 97.0 & 98.0 & 99.0 & 99.0 & 97.5 & 100.0 \\
Gemini     & 100.0 & 100.0 & 100.0 & 100.0 & 100.0 & 99.5 & 100.0 & 100.0 & 100.0 & 100.0 & 100.0 & 100.0 & 100.0 & 100.0 & 100.0 & 100.0 \\
GLM        & 100.0 & 99.0 & 94.0 & 93.0 & 91.5 & 91.0 & 95.0 & 96.5 & 93.5 & 91.5 & 92.0 & 89.0 & 91.0 & 91.0 & 94.0 & 86.0 \\
GPT        & 100.0 & 100.0 & 100.0 & 100.0 & 99.0 & 99.5 & 99.5 & 100.0 & 99.5 & 100.0 & 99.0 & 98.5 & 99.5 & 100.0 & 98.5 & 96.0 \\
Llama      & 96.0 & 96.0 & 93.0 & 96.0 & 91.0 & 88.0 & 92.0 & 92.0 & 94.0 & 94.0 & 94.0 & 89.0 & 94.0 & 99.0 & 99.0 & 98.0 \\
Wizard     & 85.5 & 42.5 & 31.5 & 20.5 & 3.0 & 38.0 & 36.5 & 23.0 & 12.0 & 13.5 & 3.5 & 5.0 & 1.5 & 8.5 & 24.5 & 90.0 \\
Qwen 7b   & 85.5 & 93.5 & 98.0 & 99.5 & 98.5 & 93.0 & 98.0 & 99.5 & 96.0 & 70.5 & 45.0 & 70.5 & 64.0 & 74.0 & 81.0 & 87.5 \\
Qwen 14b  & 93.5 & 80.5 & 93.0 & 93.5 & 98.5 & 93.0 & 93.5 & 98.0 & 96.0 & 94.5 & 96.0 & 96.0 & 97.0 & 98.5 & 96.0 & 98.5 \\
Qwen 32b  & 98.0 & 98.0 & 99.0 & 99.0 & 99.5 & 89.0 & 98.0 & 98.0 & 97.5 & 97.0 & 98.5 & 95.5 & 93.9 & 96.0 & 94.5 & 93.5 \\
Qwen 72b  & 99.5 & 99.5 & 98.0 & 96.0 & 92.5 & 93.5 & 96.5 & 98.0 & 99.0 & 99.5 & 99.5 & 100.0 & 99.5 & 100.0 & 100.0 & 100.0 \\
\bottomrule[1.1pt]
\end{tabular}
}
\caption{Performance of various models across different \textbf{absolute position} levels in \textit{Tabel SQL}. The model names are abbreviated for better layout. Full names are listed in Section~\ref{models}.}
\label{tab:abs}
\end{table*}

\begin{table*}[htbp]
\centering
\small
\scalebox{0.78}{
\begin{tabular}{lcccccccccccccccc}
\toprule[1.1pt]
\multirow{2}{*}{\vspace{-0.65em}\textbf{Model}} & \multicolumn{16}{c}{\textbf{Performance / \%}} \\
\cmidrule(lr){2-17}
 & Lv. 1 & Lv. 2 & Lv. 3 & Lv. 4 & Lv. 5 & Lv. 6 & Lv. 7 & Lv. 8 & Lv. 9 & Lv. 10 & Lv. 11 & Lv. 12 & Lv. 13 & Lv. 14 & Lv. 15 & Lv. 16 \\
\midrule[0.7pt]
Claude     & 100.0 & 74.0 & 68.0 & 65.5 & 72.5 & 63.0 & 71.0 & 60.5 & 63.0 & 61.0 & 66.0 & 64.0 & 65.0 & 67.5 & 66.5 & 64.0 \\
Deepseek   & 100.0 & 78.0 & 81.0 & 82.0 & 82.0 & 79.5 & 69.0 & 81.0 & 72.0 & 79.0 & 70.5 & 69.0 & 75.0 & 78.0 & 76.0 & 81.5 \\
Gemini     & 100.0 & 97.5 & 89.0 & 81.0 & 78.0 & 78.5 & 84.5 & 79.0 & 79.5 & 78.5 & 79.5 & 74.0 & 77.5 & 74.0 & 75.0 & 82.5 \\
GLM        & 90.0 & 69.0 & 68.5 & 67.5 & 63.0 & 58.0 & 65.0 & 48.5 & 62.0 & 50.5 & 60.0 & 57.5 & 61.5 & 52.0 & 51.5 & 44.0 \\
GPT        & 100.0 & 84.5 & 86.5 & 82.5 & 70.5 & 74.0 & 86.5 & 80.0 & 83.0 & 76.5 & 81.5 & 78.0 & 78.5 & 77.0 & 73.5 & 80.0 \\
Llama      & 100.0 & 77.0 & 77.0 & 75.0 & 88.0 & 75.0 & 79.0 & 74.0 & 74.0 & 80.0 & 68.0 & 66.0 & 75.0 & 72.0 & 79.0 & 69.0 \\
Wizard     & 74.0 & 28.5 & 23.5 & 22.5 & 47.5 & 47.0 & 61.5 & 51.5 & 54.0 & 56.5 & 65.5 & 61.0 & 61.5 & 60.0 & 61.0 & 59.5 \\
Qwen 7b    & 95.0 & 39.0 & 42.5 & 53.0 & 61.0 & 56.0 & 39.0 & 48.5 & 42.0 & 51.5 & 36.0 & 48.0 & 36.0 & 40.0 & 45.5 & 42.5 \\
Qwen 14b   & 99.5 & 59.0 & 59.0 & 63.0 & 68.5 & 56.0 & 58.5 & 55.0 & 59.5 & 59.0 & 58.0 & 62.5 & 59.5 & 62.5 & 54.0 & 63.0 \\
Qwen 32b   & 99.5 & 72.0 & 69.0 & 65.5 & 75.5 & 68.0 & 60.5 & 71.0 & 64.0 & 61.5 & 66.0 & 64.0 & 67.5 & 64.0 & 64.0 & 69.0 \\
Qwen 72b   & 100.0 & 81.0 & 77.0 & 85.0 & 87.5 & 72.5 & 67.5 & 63.5 & 67.0 & 66.0 & 67.5 & 73.5 & 70.0 & 75.5 & 75.5 & 83.0 \\
\bottomrule[1.1pt]
\end{tabular}
}
\caption{Performance of various models across different \textbf{relative position} levels in \textit{Tabel SQL}. The model names are abbreviated for better layout. Full names are listed in Section~\ref{models}.}
\label{tab:rlt}
\end{table*}

\begin{table*}[htbp]
\centering
\small
\scalebox{0.78}{
\begin{tabular}{lcccccccccccccccc}
\toprule[1.1pt]
\multirow{2}{*}{\vspace{-0.65em}\textbf{Model}} & \multicolumn{16}{c}{\textbf{Absolute Performance / \%}} \\
\cmidrule(lr){2-17}
 & Lv. 1 & Lv. 2 & Lv. 3 & Lv. 4 & Lv. 5 & Lv. 6 & Lv. 7 & Lv. 8 & Lv. 9 & Lv. 10 & Lv. 11 & Lv. 12 & Lv. 13 & Lv. 14 & Lv. 15 & Lv. 16 \\
\midrule[0.7pt]
\textbf{Claude} & 54.28 & 55.19 & 52.55 & 51.44 & 62.81 & 65.84 & 56.86 & 46.59 & 60.76 & 65.84 & 56.86 & 65.84 & 60.41 & 62.81 & 60.81 & 56.40 \\
\textbf{Gemini} & 81.37 & 76.37 & 75.59 & 73.87 & 83.34 & 81.81 & 83.46 & 83.95 & 84.55 & 83.95 & 84.15 & 84.78 & 82.81 & 84.55 & 81.94 & 80.31 \\
\textbf{GPT} & 47.64 & 57.34 & 44.99 & 43.00 & 44.57 & 53.65 & 48.37 & 50.20 & 50.78 & 48.66 & 47.45 & 49.87 & 49.37 & 54.72 & 60.61 & 45.37 \\
\bottomrule[1.1pt]
\end{tabular}
}
\caption{Performance of various models across different \textbf{absolute levels} in \textit{Code Completion}. The data includes \textbf{absolute} scores for the Claude, Gemini, and GPT models.}
\label{tab:absolute_performance2}
\end{table*}

\begin{table*}[htbp]
\centering
\small
\scalebox{0.78}{
\begin{tabular}{lcccccccccccccccc}
\toprule[1.1pt]
\multirow{2}{*}{\vspace{-0.65em}\textbf{Model}} & \multicolumn{16}{c}{\textbf{Relative Performance / \%}} \\
\cmidrule(lr){2-17}
 & Lv. 1 & Lv. 2 & Lv. 3 & Lv. 4 & Lv. 5 & Lv. 6 & Lv. 7 & Lv. 8 & Lv. 9 & Lv. 10 & Lv. 11 & Lv. 12 & Lv. 13 & Lv. 14 & Lv. 15 & Lv. 16 \\
\midrule[0.7pt]
\textbf{Claude} & 67.50 & 57.27 & 56.64 & 52.69 & 53.18 & 60.16 & 47.24 & 43.26 & 64.58 & 52.96 & 47.43 & 62.41 & 47.01 & 51.31 & 48.79 & 52.96 \\
\textbf{Gemini} & 83.87 & 83.65 & 74.71 & 84.36 & 84.36 & 81.37 & 84.11 & 74.29 & 83.09 & 75.45 & 83.09 & 75.59 & 84.36 & 77.47 & 76.45 & 72.61 \\
\textbf{GPT} & 54.57 & 51.13 & 56.03 & 58.11 & 52.19 & 48.57 & 41.59 & 51.64 & 68.87 & 62.38 & 44.21 & 42.66 & 51.66 & 45.25 & 66.06 & 54.83 \\
\bottomrule[1.1pt]
\end{tabular}
}
\caption{Performance of various models across different \textbf{relative levels} in \textit{Code Completion}. The data includes \textbf{relative} scores for the Claude, Gemini, and GPT models. Code Completion!}
\label{tab:relative_performance2}
\end{table*}

\begin{table*}[htbp]
\centering
\small
\scalebox{0.78}{
\begin{tabular}{lcccccccccccccccc}
\toprule[1.1pt]
\multirow{2}{*}{\vspace{-0.65em}\textbf{Model}} & \multicolumn{16}{c}{\textbf{Absolute Performance / \%}} \\
\cmidrule(lr){2-17}
 & Lv. 1 & Lv. 2 & Lv. 3 & Lv. 4 & Lv. 5 & Lv. 6 & Lv. 7 & Lv. 8 & Lv. 9 & Lv. 10 & Lv. 11 & Lv. 12 & Lv. 13 & Lv. 14 & Lv. 15 & Lv. 16 \\
\midrule[0.7pt]
\textbf{Claude} & 96.33 & 98.33 & 99.00 & 95.00 & 93.33 & 95.00 & 96.00 & 95.67 & 97.67 & 97.00 & 99.00 & 94.33 & 93.00 & 91.33 & 92.33 & 92.00 \\
\textbf{Gemini} & 98.00 & 98.00 & 98.00 & 98.00 & 98.00 & 97.00 & 98.00 & 97.67 & 98.00 & 98.00 & 97.33 & 98.00 & 98.00 & 98.00 & 95.33 & 98.00 \\
\textbf{GPT} & 100.00 & 99.00 & 100.00 & 98.00 & 98.00 & 100.00 & 99.00 & 100.00 & 96.00 & 97.00 & 98.00 & 99.00 & 99.00 & 98.00 & 96.00 & 96.00 \\
\bottomrule[1.1pt]
\end{tabular}
}
\caption{Performance of various models across different \textbf{absolute levels} in \textit{Wiki Retrieval}. The data includes \textbf{absolute} scores for the Claude, Gemini, and GPT models.}
\label{tab:wiki_absolute_performance}
\end{table*}

\begin{table*}[htbp]
\centering
\small
\scalebox{0.78}{
\begin{tabular}{lcccccccccccccccc}
\toprule[1.1pt]
\multirow{2}{*}{\vspace{-0.65em}\textbf{Model}} & \multicolumn{16}{c}{\textbf{Relative Performance / \%}} \\
\cmidrule(lr){2-17}
 & Lv. 1 & Lv. 2 & Lv. 3 & Lv. 4 & Lv. 5 & Lv. 6 & Lv. 7 & Lv. 8 & Lv. 9 & Lv. 10 & Lv. 11 & Lv. 12 & Lv. 13 & Lv. 14 & Lv. 15 & Lv. 16 \\
\midrule[0.7pt]
\textbf{Claude} & 100.00 & 96.00 & 99.67 & 96.00 & 98.00 & 96.33 & 95.00 & 95.67 & 93.00 & 92.67 & 91.00 & 93.00 & 95.33 & 92.00 & 96.67 & 96.33 \\
\textbf{Gemini} & 99.67 & 98.00 & 95.33 & 92.00 & 93.67 & 93.33 & 94.33 & 85.67 & 94.00 & 91.00 & 93.00 & 93.00 & 94.00 & 91.67 & 95.00 & 96.00 \\
\textbf{GPT} & 100.00 & 98.00 & 96.00 & 98.00 & 100.00 & 98.00 & 100.00 & 95.00 & 95.00 & 97.00 & 96.00 & 97.00 & 97.00 & 100.00 & 97.00 & 98.00 \\
\bottomrule[1.1pt]
\end{tabular}
}
\caption{Performance of various models across different \textbf{relative levels} in \textit{Wiki Retrieval}. The data includes \textbf{relative} scores for the Claude, Gemini, and GPT models.}
\label{tab:wiki_relative_performance}
\end{table*}

\begin{table*}[htbp]
\centering
\small
\scalebox{0.7}{
\begin{tabular}{llcccccccccccccccc}
\toprule[1.1pt]
\multirow{2}{*}{\vspace{-0.65em}\textbf{Model}} & \multirow{2}{*}{\vspace{-0.65em}\textbf{Query Position}} & \multicolumn{16}{c}{\textbf{Performance / \%}} \\
\cmidrule(lr){3-18}
 &  & Lv. 1 & Lv. 2 & Lv. 3 & Lv. 4 & Lv. 5 & Lv. 6 & Lv. 7 & Lv. 8 & Lv. 9 & Lv. 10 & Lv. 11 & Lv. 12 & Lv. 13 & Lv. 14 & Lv. 15 & Lv. 16 \\
\midrule[0.7pt]
\multirow{3}{*}{GPT} & Head        & 100.0 & 90.5 & 85.0 & 89.5 & 98.0 & 99.5 & 95.0 & 100.0 & 90.0 & 95.0 & 89.0 & 83.0 & 100.0 & 83.5 & 69.0 & 86.5 \\
 & Tail        & 100.0 & 80.0 & 36.0 & 47.0 & 68.0 & 73.5 & 81.5 & 84.5 & 70.5 & 81.5 & 79.5 & 60.0 & 68.0 & 72.5 & 67.0 & 83.0 \\
 & Both        & 100.0 & 100.0 & 100.0 & 100.0 & 99.0 & 99.5 & 99.5 & 100.0 & 99.5 & 100.0 & 99.0 & 98.5 & 99.5 & 100.0 & 98.5 & 96.0 \\
\midrule[0.7pt]
\multirow{3}{*}{Qwen 14B} & Head   & 93.5 & 84.5 & 91.0 & 96.0 & 97.5 & 90.5 & 96.5 & 97.5 & 95.0 & 93.5 & 94.0 & 95.0 & 96.5 & 98.5 & 98.5 & 97.5 \\
 & Tail   & 82.5 & 57.0 & 72.5 & 88.5 & 88.0 & 79.0 & 86.0 & 77.5 & 89.5 & 90.0 & 88.0 & 89.5 & 92.5 & 96.5 & 95.0 & 97.5 \\
 & Both   & 93.5 & 80.5 & 93.0 & 93.5 & 98.5 & 93.0 & 93.5 & 98.0 & 96.0 & 94.5 & 96.0 & 96.0 & 97.0 & 98.5 & 96.0 & 98.5 \\
\bottomrule[1.1pt]
\end{tabular}
}
\caption{Performance of GPT-4o-mini~\citep{openai2024gpt4o} and Qwen-2.5 14B~\citep{qwen2.5} across different \textbf{absolute position} levels with varying placement of the query. The query position can be at the head, tail, or both positions in the input.}
\label{tab:query_abs}
\end{table*}

\begin{table*}[htbp]
\centering
\small
\scalebox{0.7}{
\begin{tabular}{llcccccccccccccccc}
\toprule[1.1pt]
\multirow{2}{*}{\vspace{-0.65em}\textbf{Model}} & \multirow{2}{*}{\vspace{-0.65em}\textbf{Query Position}} & \multicolumn{16}{c}{\textbf{Performance / \%}} \\
\cmidrule(lr){3-18}
 &  & Lv. 1 & Lv. 2 & Lv. 3 & Lv. 4 & Lv. 5 & Lv. 6 & Lv. 7 & Lv. 8 & Lv. 9 & Lv. 10 & Lv. 11 & Lv. 12 & Lv. 13 & Lv. 14 & Lv. 15 & Lv. 16 \\
\midrule[0.7pt]
\multirow{3}{*}{GPT} & Head        & 95.0 & 60.0 & 68.5 & 69.5 & 60.5 & 63.0 & 64.0 & 75.5 & 54.5 & 63.5 & 66.0 & 62.0 & 67.0 & 46.5 & 60.0 & 79.0 \\
 & Tail        & 94.0 & 67.5 & 60.0 & 55.5 & 40.5 & 50.0 & 69.5 & 58.0 & 52.5 & 49.0 & 55.0 & 51.0 & 52.5 & 58.0 & 49.0 & 68.0 \\
 & Both        & 100.0 & 84.5 & 86.5 & 82.5 & 70.5 & 74.0 & 86.5 & 80.0 & 83.0 & 76.5 & 81.5 & 78.0 & 78.5 & 77.0 & 73.5 & 80.0 \\
\midrule[0.7pt]
\multirow{3}{*}{Qwen 14b} & Head   & 95.0 & 60.0 & 68.5 & 69.5 & 60.5 & 63.0 & 64.0 & 75.5 & 54.5 & 63.5 & 66.0 & 62.0 & 67.0 & 46.5 & 60.0 & 79.0 \\
 & Tail   & 94.0 & 67.5 & 60.0 & 55.5 & 40.5 & 50.0 & 69.5 & 58.0 & 52.5 & 49.0 & 55.0 & 51.0 & 52.5 & 58.0 & 49.0 & 68.0 \\
 & Both   & 100.0 & 84.5 & 86.5 & 82.5 & 70.5 & 74.0 & 86.5 & 80.0 & 83.0 & 76.5 & 81.5 & 78.0 & 78.5 & 77.0 & 73.5 & 80.0 \\
\bottomrule[1.1pt]
\end{tabular}
}
\caption{Performance of GPT and Qwen 14b across different \textbf{relative levels} with varying placement of the query. The query position can be at the head, tail, or both positions in the input.}
\label{tab:query_rlt}
\end{table*}

\end{document}